\documentclass[11pt,a4paper]{article}
\usepackage[hyperref]{aacl-ijcnlp2020}
\usepackage{times}
\usepackage{latexsym}

\usepackage{url}
\usepackage{latexsym}
\usepackage{multirow}
\usepackage{amsmath}
\usepackage{adjustbox}
\usepackage{array}
\usepackage{subfig}
\usepackage{graphicx}
\usepackage{enumitem}
\usepackage{booktabs}
\usepackage{xspace}
\newcolumntype{L}[1]{>{\raggedright\let\newline\\\arraybackslash\hspace{0pt}}m{#1}}
\newcolumntype{C}[1]{>{\centering\let\newline\\\arraybackslash\hspace{0pt}}m{#1}}
\newcolumntype{R}[1]{>{\raggedleft\let\newline\\\arraybackslash\hspace{0pt}}m{#1}}

\newcommand{\secref}[2][]{Section#1~\ref{sec:#2}}

\newcommand{\tabref}[2][]{Table#1~\ref{tab:#2}}
\newcommand{\figref}[2][]{Figure#1~\ref{fig:#2}}

\newcommand{\method}[1]{\textsc{#1}\xspace}
\newcommand{\ptgen}{\method{PTGen}}
\newcommand{\ptgenCov}{\method{PTGen+Cov}}
\newcommand{\bertext}{\method{BertExt}}
\newcommand{\bertabs}{\method{BertAbs}}
\newcommand{\bertextabs}{\method{BertExtAbs}}
\newcommand{\leadOne}{\method{Lead-1}}
\newcommand{\leadTwo}{\method{Lead-2}}
\newcommand{\leadThree}{\method{Lead-3}}
\newcommand{\lead}{\method{Lead-$N$}}
\newcommand{\oracle}{\method{Oracle}}
\newcommand{\bertscore}{\method{BERTScore}}

\newcommand{\z}{\ensuremath{\phantom{0}}}

\usepackage{microtype}

\aclfinalcopy 

\title{Liputan6: A Large-scale Indonesian Dataset for Text 
	Summarization}

\author{Fajri Koto \qquad Jey Han Lau \qquad Timothy Baldwin\\
	School of Computing and Information Systems \\
	The University of Melbourne \\
	\texttt{\small ffajri@student.unimelb.edu.au, jeyhan.lau@gmail.com, 
		tbaldwin@unimelb.edu.au} \\
}

\date{}

\begin{document}
\maketitle
\begin{abstract}
  In this paper, we introduce a large-scale Indonesian summarization
dataset. We harvest articles from \textit{Liputan6.com}, an online
news portal, and obtain 215,827 document--summary pairs. We leverage
pre-trained language models to develop benchmark extractive and
abstractive summarization methods over the dataset  with multilingual and monolingual BERT-based models. We
include a thorough error analysis by examining machine-generated
summaries that have low ROUGE scores, and expose both issues with ROUGE
itself, as well as with extractive and abstractive summarization models.
\end{abstract}

\section{Introduction}
\label{intro}

Despite having the fourth largest speaker population in the world, with
200 million native
speakers,\footnote{\url{https://www.visualcapitalist.com/100-most-spoken-languages/}.}
Indonesian is under-represented in NLP.  One reason is the scarcity of
large datasets for different tasks, such as parsing, text classification,
and summarization.  In this paper, we attempt to bridge this gap by
introducing a large-scale Indonesian corpus for text summarization.

Neural models have driven remarkable progress in summarization in recent
years, particularly for abstractive summarization. One of the first
studies was \newcite{rush2015neural}, where the authors proposed an
encoder--decoder model with attention to generate headlines for English
Gigaword documents \cite{graff2003english}.  Subsequent studies
introduced pointer networks \cite{nallapati2016abstractive,see2017get},
summarization with content selection
\cite{hsu2018unified,gehrmann2018bottom}, graph-based attentional models
\cite{tan2017abstractive}, and deep reinforcement learning
\cite{paulus2017a}. More recently, we have seen the widespread adoption
of pre-trained neural language models for summarization, e.g.\ BERT
\cite{liu2019text}, BART \cite{lewis2019bart}, and PEGASUS
\cite{zhang2019pegasus}.

Progress in summarization research has been driven by the availability
of large-scale English datasets, including 320K \textit{CNN/Daily Mail}
document--summary pairs \cite{hermann2015teaching} and 100k \textit{NYT}
articles \cite{sandhaus2008the} which have been widely used in
abstractive summarization research
\cite{see2017get,gehrmann2018bottom,paulus2017a,lewis2019bart,zhang2019pegasus}.
News articles are a natural candidate for summarization datasets, as
they tend to be well-structured and are available in large volumes.
More recently, English summarization datasets in other flavours/domains
have been developed, e.g.\ \textit{XSum} has 226K documents with highly
abstractive summaries \cite{narayan2018don}, BIGPATENT is a
summarization dataset for the legal domain \cite{sharma2019bigpatent},
Reddit TIFU is sourced from social media \cite{kim2019abstractive}, and
\newcite{cohan2018a} proposed using scientific publications from arXiv and
PubMed for abstract summarization.

\begin{figure*}[]
	\centering
	\includegraphics[width=5.7in]{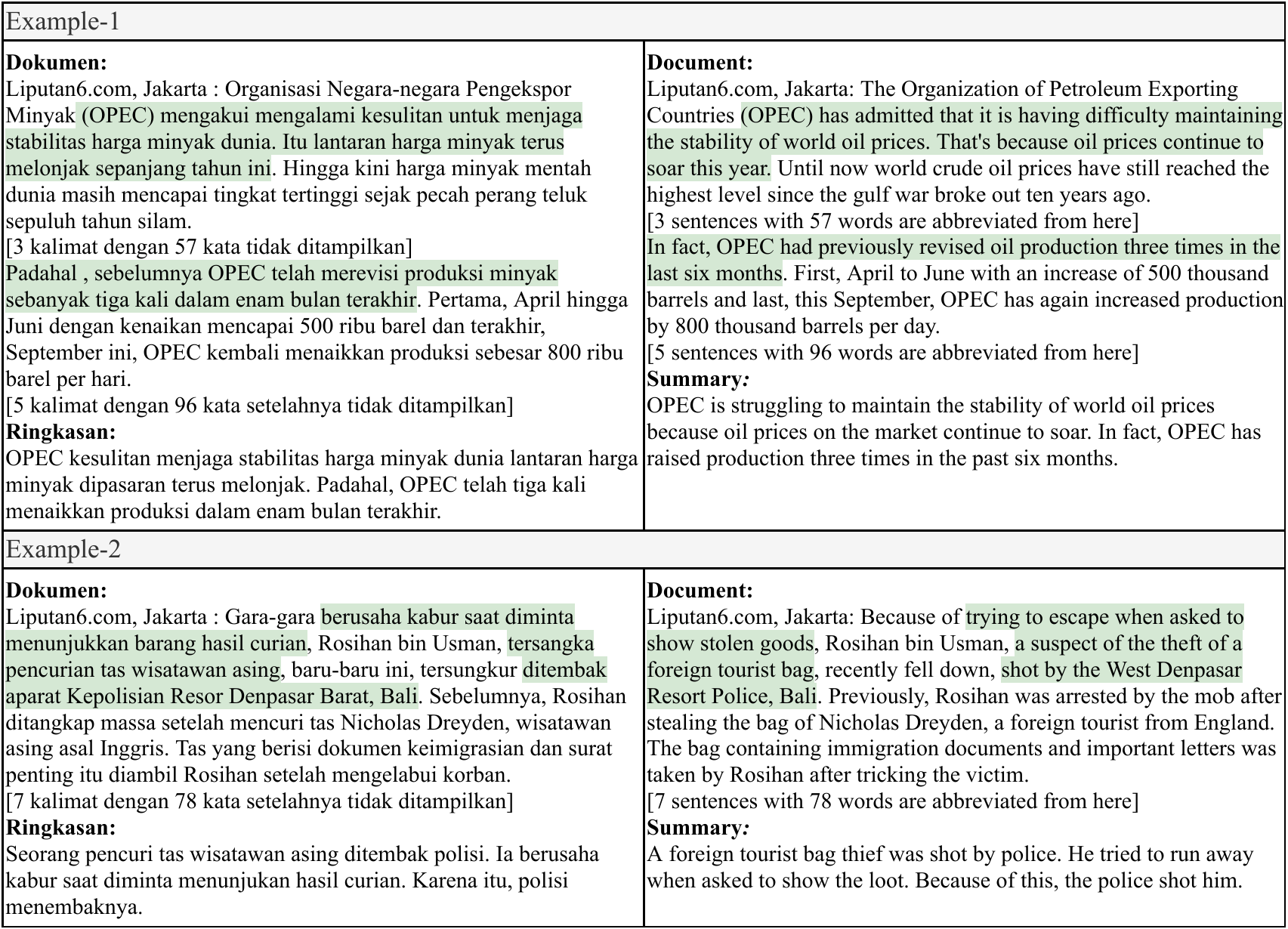}
	\caption{Example articles and summaries from Liputan6.  To the
		left is the original document and summary, and to the right is
		an English translation (for illustrative purposes).  We
		additionally highlight sentences that the summary is based
		on (noting that such highlighting is not available in the dataset).}
	\label{fig:example}
\end{figure*}

This paper introduces the first large-scale summarization 
dataset for Indonesian, sourced from the \textit{Liputan6.com} online news portal
over a 10-year period. It covers various topics and events that happened
primarily in Indonesia, from October 2000 to October 2010. Below, we
present details of the dataset, propose benchmark extractive and
abstractive summarization methods that leverage both multilingual and
monolingual pre-trained BERT models. We further conduct error analysis
to better understand the limitations of current models over the dataset,
as part of which we reveal not just modelling issues but also problems
with ROUGE.

To summarize, our contributions are: (1) we release a large-scale
Indonesian summarization corpus with over 200K documents, an order of
magnitude larger than the current largest Indonesian summarization
dataset and one of the largest non-English summarization datasets in
existence;\footnote{The data can be accessed at
  \url{https://github.com/fajri91/sum_liputan6}} (2) we present
statistics to show that the summaries in the dataset are reasonably
abstractive, and provide two test partitions, a standard test set and an
extremely abstractive test set; (3) we develop benchmark extractive and
abstractive summarization models based on pre-trained BERT models; and
(4) we conduct error analysis, on the basis of which we share insights
to drive future research on Indonesian text summarization.

\section{Data Construction}
\label{sec:data}

\textit{Liputan6.com} is an online Indonesian news portal which has been
running since August 2000, and provides news across a wide range of
topics including politics, business, sport, technology, health, and
entertainment. According to the Alexa ranking of websites at the time of
writing,\footnote{\url{https://www.alexa.com/topsites}} \textit{Liputan6.com}
is ranked 9th in Indonesia and 112th globally.  The website produces
daily articles along with a short description for its RSS feed.  The
summary is encapsulated in the javascript variable
\texttt{window.kmklabs.article} and the key \texttt{shortDescription},
while the article is in the main body of the associated HTML page. We
harvest this data over a 10-year window --- from October 2000 to October
2010 --- to create a large-scale summarization corpus, comprising
215,827 document--summary pairs. In terms of preprocessing, we remove
formatting and HTML entities (e.g. \texttt{\&quot}, and \texttt{\_\_}),
lowercase all words, and segment sentences based on simple punctuation
heuristics.  We provide example articles and summaries, with English
translations for expository purposes (noting that translations are
not part of the dataset), in \figref{example}.

\begin{table}[t]
	\begin{center}
		\begin{adjustbox}{max width=1\linewidth}
		\begin{tabular}{cr@{\;\;}r@{\;\;}rr@{\;\;}r@{\;\;}r@{\;\;}r}
			\toprule
			\multirow{2}{*}{\bf Variant} & \multicolumn{3}{c}{\bf 
				\#Doc} & \multicolumn{4}{c}{\bf \% of Novel $n$-grams} 
			\\
			& \bf Train & \bf Dev & \bf Test & \bf 1 & \bf 2 & \bf 3 
			& \bf 
			4 \\
			\midrule
			Canonical & 193,883 & 10,972 & 10,972 & 16.2 & 52.5 & 
			71.8 & 
			82.4 \\
			Xtreme & 193,883 & 4,948 & 3,862 & 22.2 & 66.7 & 
			87.5 & 
			96.6 \\
			\bottomrule
		\end{tabular}
		\end{adjustbox}
	\end{center}
	\caption{Statistics for the canonical and Xtreme variants of our
          data. The percentage of novel \textit{n}-grams is based on the
          combined Dev and Test set.}
	\label{tab:valtest}
\end{table}

\begin{figure}[t]
	\centering
	\includegraphics[width=3.1in]{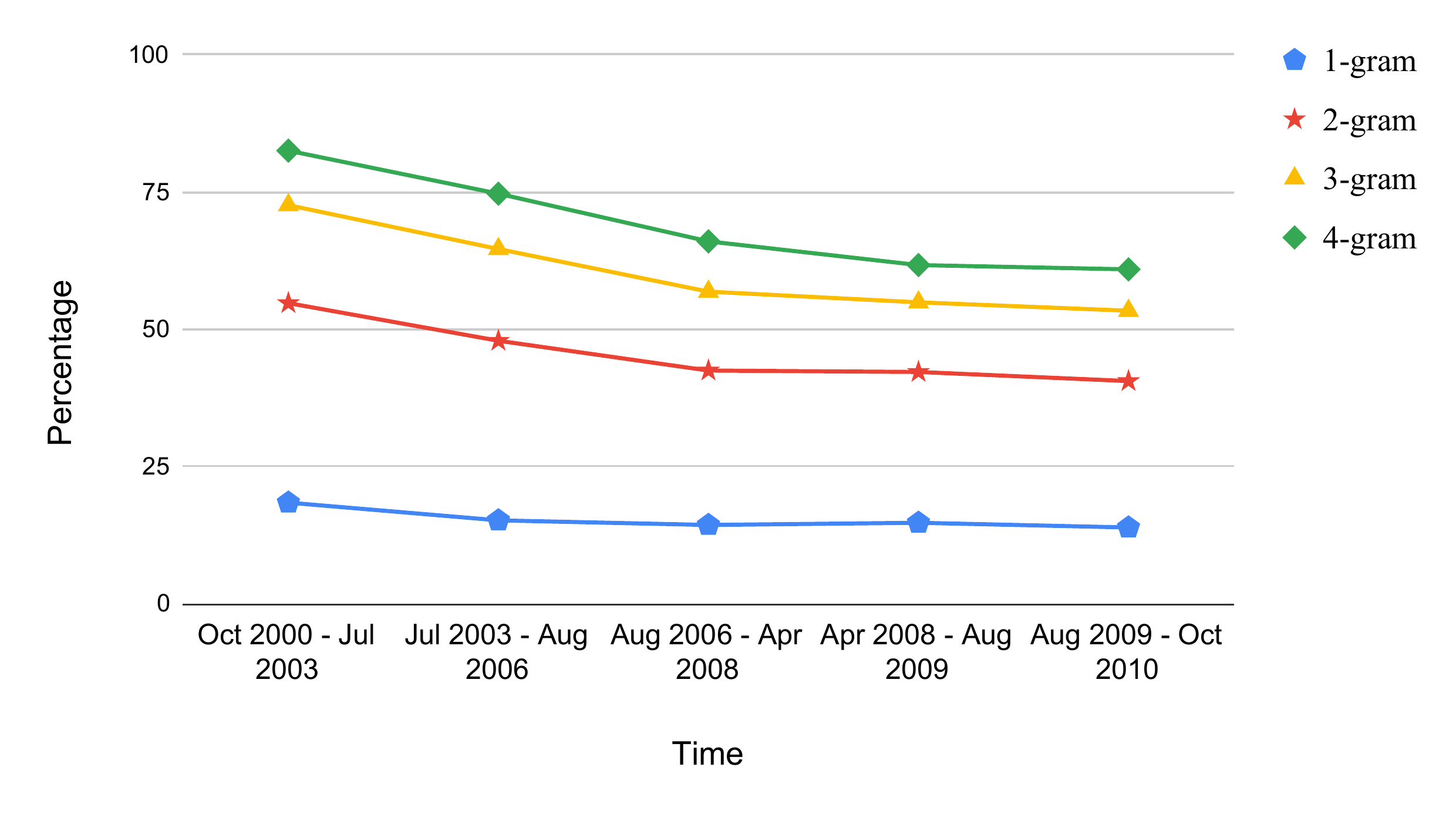}
	\caption{Proportion of novel $n$-grams over time in the
		summaries.}
	\label{fig:abs}
\end{figure}

As a preliminary analysis of the document--summary pairs over the
10-year period, we binned the pairs into 5 chronologically-ordered
groups containing 20\% of the data each, and computed the proportion of
novel $n$-grams (order 1 to 4) in the summary (relative to the source document). 
Based on the results in \figref{abs}, we can see that the proportion of
novel $n$-grams drops over time, implying that the summaries of more
recent articles are less abstractive.  For this reason, we decide to use
the earlier articles (October 2000 to Jan 2002) as the development and
test documents, to create a more challenging dataset.  This setup also
means there is less topic overlap between training and development/test
documents, allowing us to assess whether the summarization models are
able to summarize unseen topics.

\begin{table*}[t]
	\begin{center}
		\begin{adjustbox}{max width=0.85\linewidth}
			\begin{tabular}{cr@{\;}r@{\;}rc@{\;}c@{\;}cc@{\;}c@{\;}c}
				\toprule
				\multirow{2}{*}{\bf Dataset} & \multicolumn{3}{c}{\bf \#Doc} 
				& \multicolumn{3}{c}{\bf Article} & \multicolumn{3}{c}{\bf Summary}  \\
				& \bf Train & \bf Dev & \bf Test  & \bf $\mu$(Word)& \bf 
				$\mu$(Sent) & \bf \#Vocab & \bf $\mu$(Word) & \bf $\mu$(Sent) & \bf 
				\#Vocab \\
				\midrule
				IndoSum & 14,252 & 750 & 3,762 & 347.23 & 18.37 &117K & 
				68.09 & 3.47 & 53K \\
				Liputan6 & 193,883 & 10,972 & 10,972 & 232.91 & 12.60 & 311K 
				& 30.43 & 2.09  & 100K \\
				\bottomrule
			\end{tabular}
		\end{adjustbox}
	\end{center}
	\caption{\label{tab:datastat} A comparison of IndoSum and Liputan6.
		$\mu$(Word) and $\mu$(Sent) denote the average number of words and
		sentences, respectively.}
\end{table*}

\begin{table}[t]
	\begin{center}
		\begin{adjustbox}{max width=\linewidth}
			\begin{tabular}{cr@{\;\;}r@{\;\;}rr@{\;\;}r@{\;\;}r@{\;\;}r}
				\toprule
				\multirow{2}{*}{\bf Dataset} & \multicolumn{3}{c}{\bf 
					\lead} & \multicolumn{4}{c}{\bf \% of Novel $n$-grams} \\
				& \bf R1 & \bf R2 & \bf RL & \bf 1 & \bf 2 & \bf 3 & \bf 4 \\
				\midrule
				IndoSum & 65.6 & 58.9 & 64.8 & 3.1 & 10.8 & 16.2 & 
				20.3 \\
				Liputan6 & 41.2 & 27.1 & 38.7 & 12.9 & 41.6 & 57.6 & 
				66.9\\
				\bottomrule
			\end{tabular}
		\end{adjustbox}
	\end{center}
	\caption{\label{tab:abs} Abstractiveness of the summaries in IndoSum 
		and Liputan6.}
\end{table}

For the training, development and test partitions, we use a splitting
ratio of 90:5:5. In addition to this canonical partitioning of the data,
we provide an ``Xtreme'' variant (inspired by \textit{Xsum};
\newcite{narayan2018don}) whereby we discard development and test
document--summary pairs where the summary has fewer than 90\% novel
$4$-grams (leaving the training data unchanged), creating a smaller,
more challenging data configuration. Summary statistics for the
``canonical'' and ``Xtreme'' variants are given in \tabref{valtest}.

We next present a comparison of Liputan6 (canonical partitioning) and
IndoSum (the current largest Indonesian summarization dataset, as
detailed in \secref{related}; \newcite{kurniawan2018indosum}) in
\tabref{datastat}.  In terms of number of documents, Liputan6 is
approximately 11 times larger than IndoSum (the current largest
Indonesian summarization dataset), although articles and summaries in Liputan6 are slightly shorter.

To understand the abstractiveness of the summaries in the two datasets,
in \tabref{abs} we present ROUGE scores for the simple baseline of using
the first $N$ sentences as an extractive summary (``\lead''), and the
percentage of novel $n$-grams in the summary.\footnote{All statistics
	are based on the entire dataset, encompassing the training, dev, and test
	data.}  We use \leadThree{} and \leadTwo{} for IndoSum and Liputan6
respectively, based on the average number of sentences in the summaries
(\tabref{datastat}).  We see that Liputan6 has consistently lower ROUGE
scores (R1, R2, and RL) for \lead; it also has a substantially higher
proportion of novel $n$-grams.  This suggests that the summaries in
Liputan6 are more abstractive than IndoSum.

To create a ground truth for extractive summarization,  we follow 
\newcite{cheng2016neural} and \newcite{nallapati2016summarunner}  in 
greedily selecting the subset of sentences in the article that maximizes 
the ROUGE score based on the reference summary. As a result, each sentence in the 
article has a binary label to indicate whether they should be included 
as part of an extractive summary. Extractive summaries created this way 
will be referred to as ``\oracle'', to denote the upper bound  
performance of an extractive summarization system.


\section{Summarization Models}
\label{model}

\begin{figure}[t]
	\centering
	\includegraphics[width=3in]{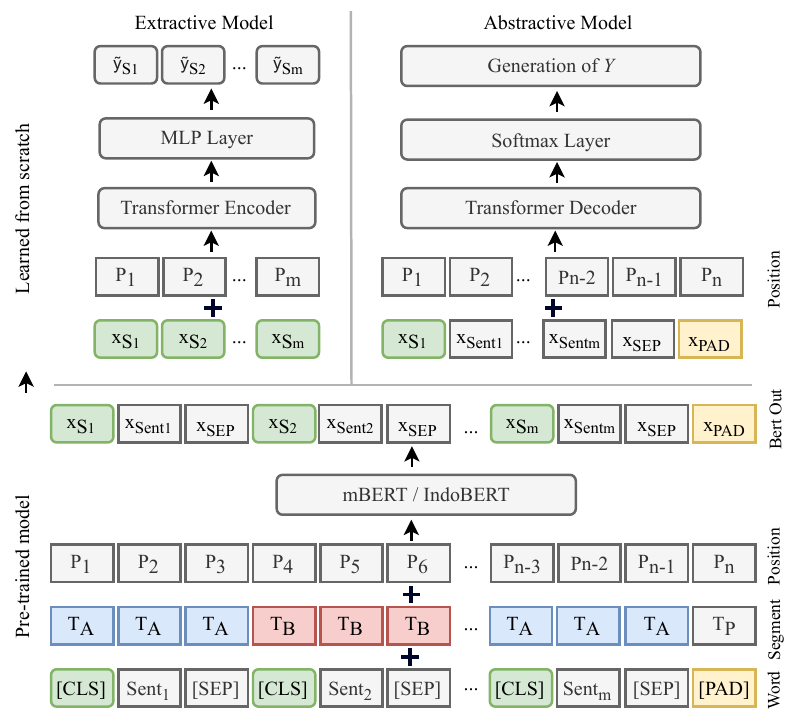}
	\caption{\label{fig:sum} Architecture of the extractive and 
		abstractive summarization models.}
\end{figure}

We follow \newcite{liu2019text} in building extractive and abstractive 
summarization models using BERT as an encoder to produce contextual 
representations for the word tokens.  The architecture of both models is 
presented in \figref{sum}.
We tokenize words with WordPiece, and append  [CLS] (prefix) and [SEP] 
(suffix) tokens to each sentence.  To further distinguish the sentences, 
we add even/odd segment embeddings ($T_A/T_B$) based on the order of the 
sentence to the word embeddings. For instance, for a document with 
sentences [$s_1, s_2, s_3, s_4$], the segment embeddings are 
[$T_A,T_B,T_A,T_B$].  Position embeddings ($P$) are also used to denote 
the position of each token.  The WordPiece, segment, and position 
embeddings are summed together and provided as input to  BERT.

BERT produces a series of contextual representations for the word 
tokens, which we feed into a (second) transformer encoder/decoder for 
the extractive/abstractive summarization model. We detail the 
architecture of these two models in \secref[s]{extractive-model} and 
\ref{sec:abstractive-model}. Note that this second transformer is 
initialized with random parameters (i.e.\ it is not pre-trained).

For the pre-trained BERT encoder, we use multilingual BERT (mBERT) and
our own IndoBERT \cite{koto2020indolem}.\footnote{The pre-trained mBERT is sourced from:
	\url{https://github.com/google-research/bert}.}  IndoBERT is a
BERT-Base model we trained ourselves using Indonesian documents from
three sources: (1) Indonesian Wikipedia (74M words); (2) news articles
(55M words) from Kompas,\footnote{\url{https://kompas.com}} Tempo
\cite{tala2003the},\footnote{\url{https://koran.tempo.co}} and
Liputan6;\footnote{For Liputan6, we use only the articles from the
	training partition.} and (3) the Indonesian Web Corpus (90M words;
\newcite{medved2019indonesian}).  In total, the training data has 220M
words. We implement IndoBERT using the Huggingface
framework,\footnote{\url{https://huggingface.co/}} and follow the
default configuration of BERT-Base (uncased): hidden size = 768d, hidden
layers = 12, attention heads = 12, and feed-forward = 3,072d.  We train IndoBERT with 31,923 WordPieces
(vocabulary) for 2 million steps.


\subsection{Extractive Model}
\label{sec:extractive-model}


After the document is processed by BERT, we have a contextualized 
embedding for every word token in the document. To learn 
inter-sentential relationships, we use the [CLS] embeddings ([$x_{S_1}, 
x_{S_2},.., x_{S_m}$]) to represent the sentences, to which we add 
a sentence-level positional embedding ($P$), and feed them to a 
transformer encoder (\figref{sum}). An MLP layer with sigmoid activation 
is applied to the output of the transformer encoder to predict whether a 
sentence should be extracted (i.e.\ $\tilde{y}_S \in \{0,1\}$).
We train the model with binary cross entropy, and update all model 
parameters (including BERT) during training.  Note that the parameters 
in the transformer encoder and the MLP layer are initialized randomly,
and learned from scratch.

The transformer encoder is configured as follows: layers = 2, 
hidden size = 768, feed-forward = 2,048, and heads = 8.
In terms of training hyper-parameters, we train using the Adam optimizer 
with learning rate 
$lr=2e^{-3}\cdot\text{min}(\text{step}^{-0.5}, \text{step}\cdot\text{warmup}^{-1.5})$ 
where $\text{warmup}=10,000$. We train for 50,000 steps on 3$\times$V100
16GB GPUs, and perform evaluation on the development set every 2,500
steps. At test time, we select sentences for the extractive summary
according to two conditions: the summary must consist of: (a) at least
two sentences, and (b) at least 15 words. These values were set based on
the average number of sentences and the minimum number of words in a
summary. We also apply trigram blocking to reduce redundancy
\cite{paulus2017a}.
Henceforth, we refer to this model as ``\bertext''.

\subsection{Abstractive Model}
\label{sec:abstractive-model}

Similar to the extractive model, we have a second transformer to process 
the contextualized embeddings from BERT. In this case, we use a 
transformer decoder instead (i.e.\ an attention mask is used to prevent the decoder from attending to future time 
steps), as we are learning to generate an abstractive summary. But 
unlike the extractive model, we use the BERT embeddings for all tokens 
as input to the transformer decoder (as we do not need sentence 
representations).  We add to these BERT embeddings a second positional 
encoding before feeding them to the transformer decoder (\figref{sum}).  
The transformer decoder is initialized with random parameters (i.e.\ no 
pre-training).

The transformer decoder is configured as follows: layers = 6, 
hidden size = 768, feed-forward = 2,048, and heads = 8.
Following \newcite{liu2019text}, we use a different learning rate for 
BERT and the decoder when training the model:  $lr$  
$=2e^{-3}\cdot\text{min}(\text{step}^{-0.5},\text{step}\cdot20,000^{-1.5})$ and 
$0.1\cdot\text{min}(\text{step}^{-0.5},\text{step}\cdot10,000^{-1.5})$ for BERT and 
the transformer decoder, respectively.  Both networks are trained with 
the Adam optimizer for 200,000 steps on 4$\times$V100 16GB GPUs and 
evaluated every 10,000 steps. For summary generation, we use beam width 
= 5, trigram blocking, and a length penalty \cite{wu2016google} to generate 
at least two sentences and at least 15 words (similar to the extractive model).

Henceforth the abstractive model will be referred to as 
``\bertabs''.  We additionally experiment with a third variant, 
``\bertextabs'', where we use the weights of the fine-tuned BERT in 
\bertext for the encoder (instead of off-the-shelf BERT 
weights).

\section{Experiment and Results}
\label{exp}

\begin{table*}[t]
	\begin{center}
		\begin{adjustbox}{max width=\linewidth}
			\begin{tabular}{ccccccccccc}
				\toprule \multirow{2}{*}{\bf Model} && 
				\multicolumn{4}{c}{\bf Canonical Test Set} && \multicolumn{4}{c}{\bf Xtreme Test Set} \\
				\cmidrule{3-6}
				\cmidrule{8-11}
				&& \bf R1 & \bf R2 & \bf RL &\bf BS && \bf R1 & \bf R2 & \bf RL 
				& \bf BS \\
				\midrule
				\leadOne && 32.67 & 18.50 & 29.40 & 72.62 && 27.27 & 11.56 & 23.60 & 71.19\\
				\leadTwo && 36.68 & 20.23 & 33.71 & 74.58 && 31.10 & 12.78 & 27.63 & 72.98\\
				\leadThree && 34.49 & 18.84 & 32.06 & 74.31 && 29.54 & 12.05 & 26.68 & 72.78\\
				\oracle && 51.54 & 30.56 & 47.75 & 79.24 && 43.69 & 18.57 & 38.84 & 76.75\\ 
				\midrule
				\ptgen && 36.10 & 19.19 & 33.56 & 75.92 && 30.41 & 12.05 & 27.51 & 74.10\\ 
				\ptgenCov && 35.53 & 18.56 & 32.92 & 75.75 && 30.27 & 11.81 & 27.26 & 74.11\\ 
				\midrule
				\bertext{} (mBERT)&& 37.51 & 20.15 & 34.57 & 75.22 && 31.83 & 12.63 & 28.37 & 73.62\\ 
				\bertabs{} (mBERT) && 39.48 & 21.59 & 36.72 & 77.19 && 33.26 & 13.82 & 30.12 & 75.40 \\ 
				\bertextabs{} (mBERT)&& 39.81 & 21.84 & 37.02 & 77.39 && 33.86 & 14.13 & 30.73 & 75.69 \\ 
				\midrule
				\bertext{} (IndoBERT)&& 38.03 & 20.72 & 35.07 & 75.33 && 31.95 & 12.74 & 28.47 & 73.64\\ 
				\bertabs{} (IndoBERT) && 40.94 & \bf 23.01 & 37.89 & 77.90 && 34.59 & \bf 15.10 & 31.19 & 75.84\\ 
				\bertextabs{} (IndoBERT)&& \bf 41.08 & 22.85 & \bf 38.01 & \bf 77.93 && \bf 34.84 & 15.03 & \bf 31.40 & \bf 75.99\\ 
				\bottomrule
			\end{tabular}
		\end{adjustbox}
	\end{center}
	\caption{ROUGE results for the canonical and 
		Xtreme test sets. All ROUGE (``R1'', ``R2'', and ``RL'')
                scores have a confidence interval of at most $\pm{0.3}$, as reported by the official ROUGE script. ``BS'' is \textsc{BERScore} computed with \texttt{bert-base-multilingual-cased} (layer 9), as suggested by \citet{zhang2020bertscore}.}
	\label{tab:expresult}
\end{table*}

We use three ROUGE \cite{lin2004rouge} F-1 scores as evaluation metrics:
R1 (unigram overlap), R2 (bigram overlap), and RL (longest common
subsequence overlap). In addition, we also provide \bertscore (F-1), as
has recently been used for machine translation evaluation \cite{zhang2020bertscore}.\footnote{\url{https://github.com/Tiiiger/bert_score}} We use the development set to select the best
checkpoint during training, and report the evaluation scores for the
canonical and Xtreme test sets in \tabref{expresult}.  For both test
sets, the summarization models are trained using the same training set,
but they are tuned with a different development set (see \secref{data}
for details). In addition to the BERT models, we also include two
pointer-generator models \cite{see2017get}: (1) the base model (\ptgen);
and (2) the model with coverage penalty (\ptgenCov).\footnote{We use the
	default hyper-parameter configuration recommended by the original
	authors for the pointer-generator models.}

We first look at the baseline \lead and \oracle results.  \leadTwo is
the best \lead baseline for Liputan6. This is unsurprising, given that
in \tabref{datastat}, the average summary length was 2 sentences. We
also notice there is a substantial gap between \oracle{} and \leadTwo:
12--15 points for R1 and 5--7 points for \bertscore, depending on the test set. This suggests that the
baseline of using the first few sentences as an extractive summary is
ineffective. Comparing the performance between the canonical and Xtreme
test sets, we see a substantial drop in performance for both \lead and
\oracle, highlighting the difficulty of the Xtreme test set due to its
increased abstractiveness.


For the pointer-generator models, we see little improvement when
including the coverage mechanism (\ptgenCov vs.\ \ptgen), implying that
there is minimal repetition in the output of \ptgen.  We suspect this is
due to the Liputan6 summaries being relatively short (2 sentences with 30
words on average).  A similar observation is reported by
\newcite{narayan2018don} for \textit{XSum}, where the summaries are
similarly short (a single sentence with 23 words, on average).

Next we look at the BERT models. Overall they perform very well, with
both the mBERT and IndoBERT models outperforming the \lead baselines and
\ptgen models by a comfortable margin.  IndoBERT is better than mBERT
(approximately 1 ROUGE point better on average over most metrics),
showing that a monolingually-trained BERT is a more effective
pre-trained model than the multilingual variant. The best performance is
achieved by IndoBERT's \bertextabs. In the canonical test set, the
improvement over \leadTwo is $+$4.4 R1, $+$2.62 R2, $+$4.3
R3, and $+$3.4 \bertscore points. In the Xtreme test set, \bertextabs suffers a
substantial drop compared to the canonical test set (6--7 ROUGE and 2 \bertscore points),
although the performance gap between it and \leadTwo is about the
same.

\section{Error Analysis}
\label{error}

In this section, we analyze errors made by the extractive (\bertext) and
abstractive (\bertextabs) models to better understand their
behaviour. We use the mBERT version of these models in our
analysis.\footnote{The error analysis is based on mBERT rather than
  IndoBERT simply because this was the best-performing model at the time
  the error analysis was performed. While IndoBERT ultimately performed
  slightly better, given that the two models are structurally identical,
  we would expect to see a similar pattern of results.}


\subsection{Error Analysis of Extractive Summaries}

\begin{figure*}[!t]
	\centering
	\subfloat[Distribution of sentence positions for \oracle and 
	\bertext in the canonical test 
	set.]{\includegraphics[width=0.48\textwidth]{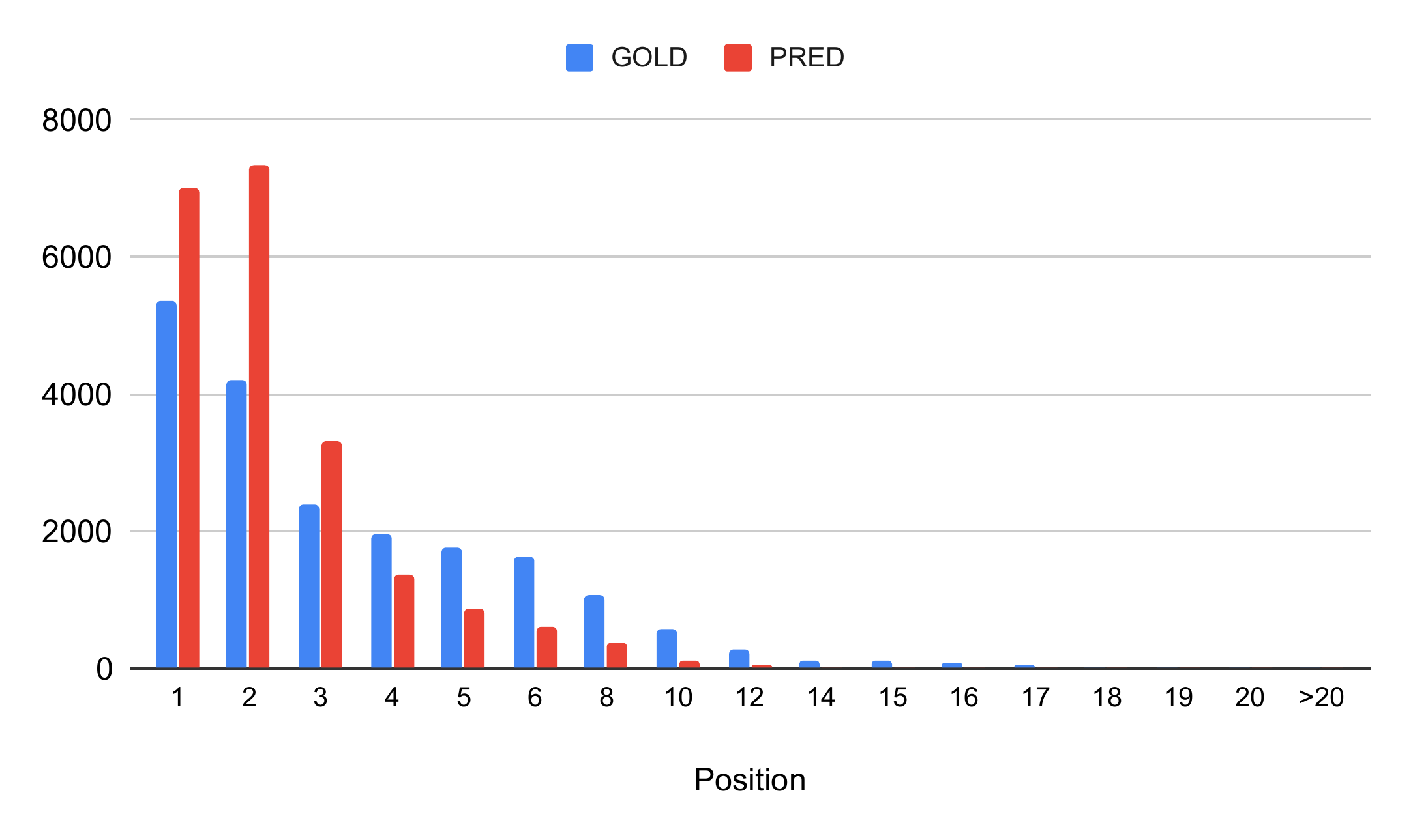}\label{fig:f1}}
	\hfill
	\subfloat[Distribution of sentence positions for \oracle in 
	the training  
	set.]{\includegraphics[width=0.48\textwidth]{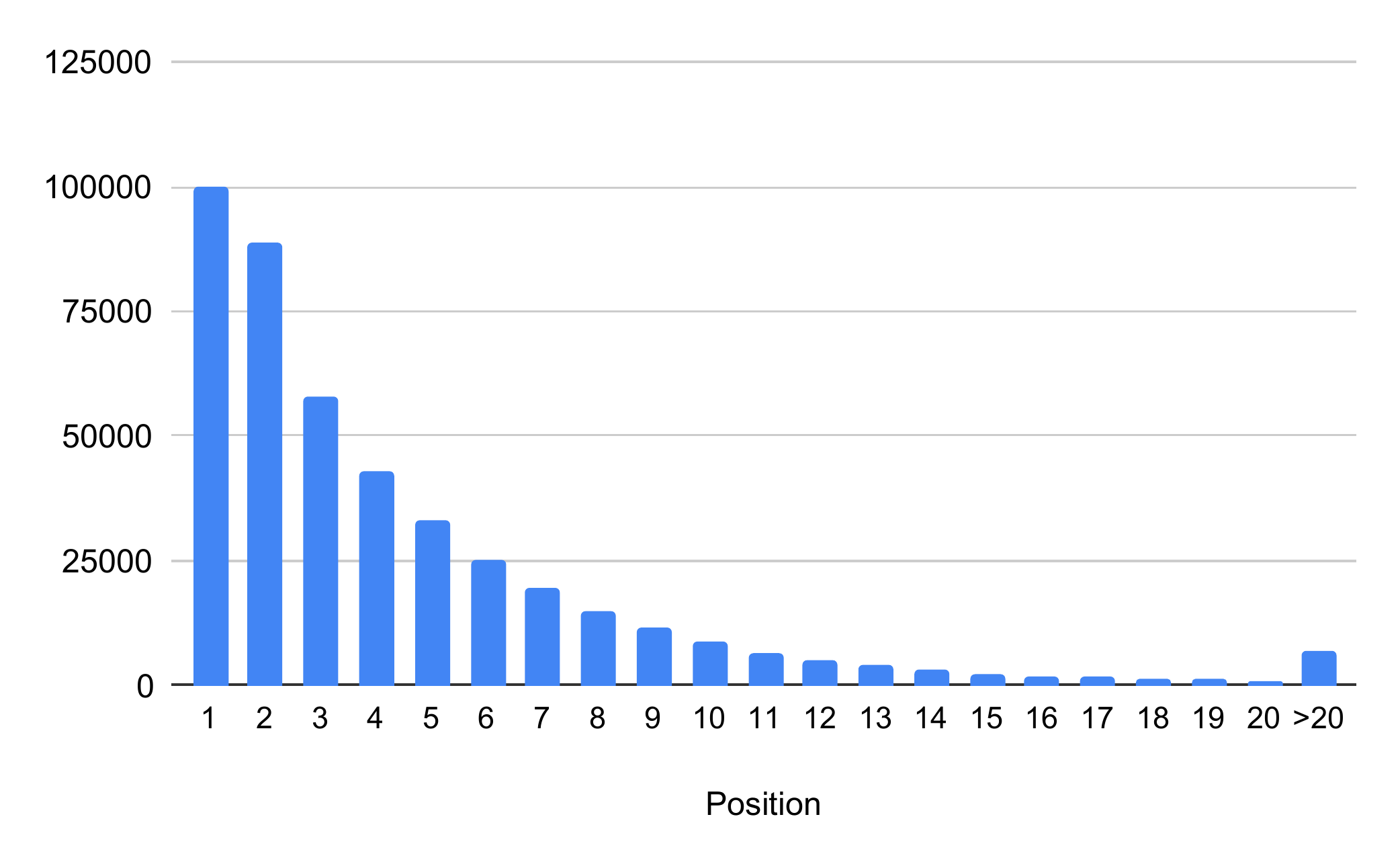}\label{fig:f2}}
	\caption{\label{fig:pos} Position of \oracle and/or Predicted Extractive Summaries}
\end{figure*}

We hypothesized that the disparity between \oracle and 
\bertext (14.03 point difference for R1 in the canonical test 
set) was due to the number of extracted sentences. To test this, when 
extracting sentences with \bertext, we set the total number of 
extracted sentences to be the same as the number of sentences in the 
\oracle summary. However, we found minimal benefit using this 
approach, suggesting that the disparity is not a result of the number of 
extracted sentences.


To investigate this further, we present the frequency of
\textit{sentence positions} that are used in the summary in \oracle and
\bertext for the canonical test set in \figref{f1}. We can see that
\bertext tends to over-select the first two sentences as the summary.
In terms of proportion, 65.47\% of \bertext summaries involve the first
two sentences. In comparison, only 42.54\% of \oracle summaries use
sentences in these positions. One may argue that this is because the
training and test data have different distributions under our chronological
partitioning strategy (recall that the test set is sampled from the
earliest articles), but that does not appear to be the case: as
\figref{f2} shows, the distribution of sentence positions in the
training data is very similar to the test data --- 43.14\% of \oracle
summaries involve the first two sentences.

\subsection{Error Analysis of Abstractive Summaries}

\begin{figure*}[t]
	\centering
	\includegraphics[width=6in]{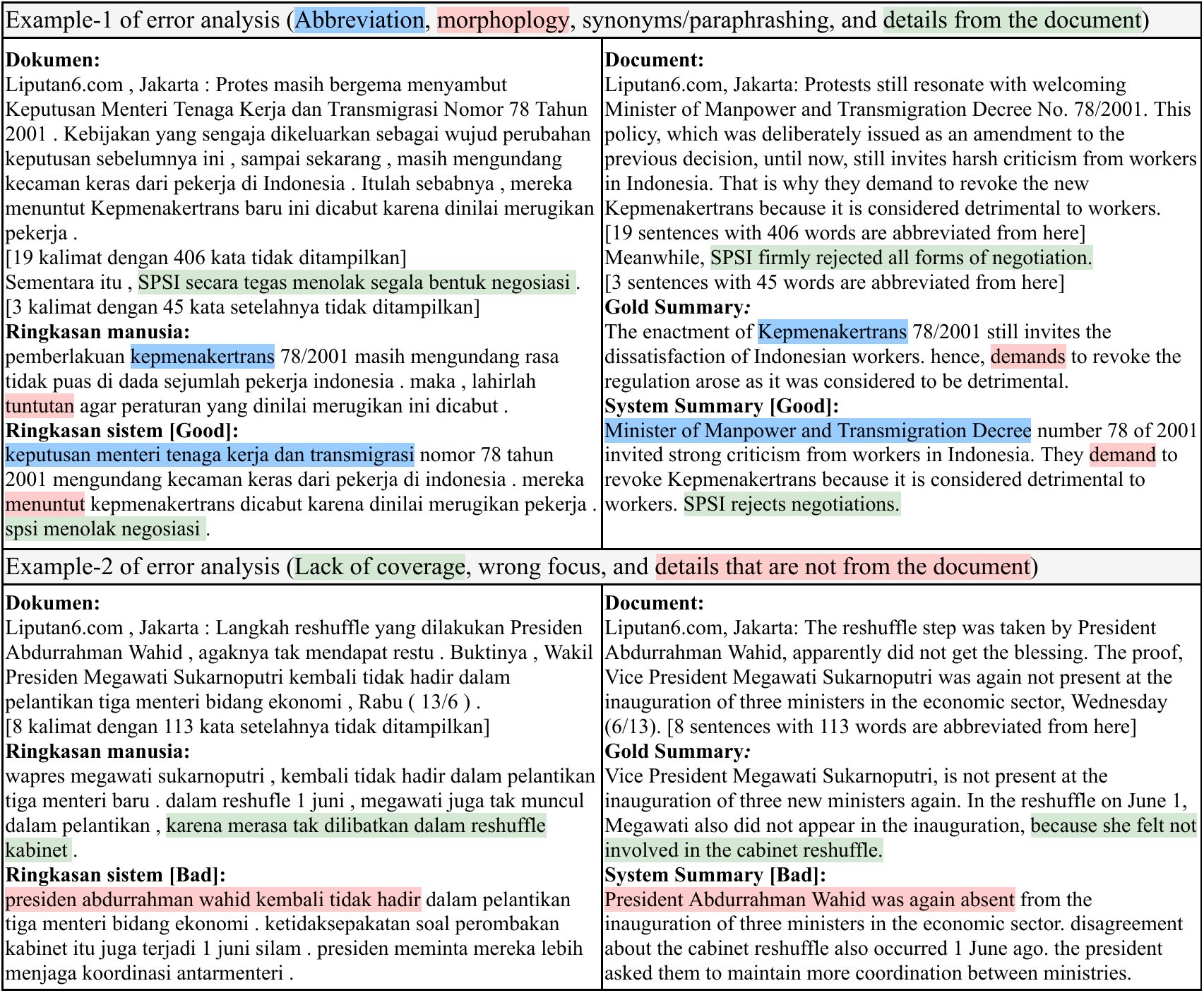}
	\caption{\label{fig:error} Two examples to highlight error categories used in our error analysis.}
\end{figure*}

To perform error analysis for \bertextabs, we randomly sample 100
documents with an R1 score $<$0.4 in the canonical test set (which
accounts for nearly 50\% of the test documents). Two native Indonesian
speakers examined these 100 samples to manually assess the quality of
the summaries, and score them on a 3-point ordinal scale: (1)
\textit{bad}; (2) \textit{average}; and (3) \textit{good}. Each
annotator is presented with the source document, the reference summary,
and the summary generated by \bertextabs. In addition to the overall
quality evaluation, we also asked the annotators to analyze a number of
(fine-grained) attributes in the summaries:


\begin{itemize}[noitemsep]
	\item \textit{Abbreviations:} the system summary uses abbreviations that
	are different to the reference summary.
	\item \textit{Morphology:} the system summary uses morphological
	variants of the same lemmas contained in the reference summary.
	\item \textit{Synonyms/paraphrasing:} the system summary contains
	paraphrases of the reference summary.
	\item \textit{Lack of coverage:} the system summary lacks coverage of
	certain details that are present in the reference summary.
	\item \textit{Wrong focus:} the system summarizes a different 
	aspect/focus of the document to the reference summary.
	\item \textit{Unnecessary details (from document):} the system
	summary includes unimportant but factually correct information.
	\item \textit{Unnecessary details (not from document):} the system
	summary includes unimportant and factually incorrect information
	(hallucinations).
\end{itemize}

\begin{table}[t!]
	\begin{center}
		\begin{adjustbox}{max width=1\linewidth}
			\begin{tabular}{rrrr}
				\toprule
				\bf Category & \bf Bad & \bf Avg. & \bf Good \\
				\midrule
				\#Samples (100) & 32\z & 8\z & 60\z\\
				Abbreviation (\%) & 21.9 & 25.0 & 40.0 \\
				Morphology (\%) & 12.5 & 25.0 & 36.7 \\
				Paraphrasing (\%) & 50.0 & 87.5 & \bf 86.7 \\
				Lack of coverage (\%) & \bf 90.6 & \bf 100.0 & 40.0 \\
				Wrong focus (\%) & 68.8 & 0.00 & 8.3 \\
				Un. details (from doc) (\%) & \bf 90.6 &  75.0 & 75.0 \\
				Un. details (not from doc) (\%) & 18.8 & 12.5 & 5.0 \\
				\bottomrule
			\end{tabular}
		\end{adjustbox}
	\end{center}
	\caption{\label{tab:err} Error analysis for 100 samples with R1
          $<$0.4.}
\end{table}

We present a breakdown of the different error types in \tabref{err}.
Inter-annotator agreement for the overall quality assessment is high
(Pearson's $r =$ 0.69). Disagreements in the quality label (\textit{bad,
  average, good}) are resolved as follows: (1) \{\textit{bad},
\textit{average}\} $\rightarrow$ \textit{bad}; and (2) \{\textit{good},
\textit{average}\} $\rightarrow$ \textit{good}. We only have four
examples with \{\textit{bad}, \textit{good}\} disagreement, which we
resolved through discussion.  Interestingly, more than half (60) of our
samples were found to have \textit{good} summaries.  The primary reasons
why these summaries have low ROUGE scores are paraphrasing (86.7\%),
and the inclusion of additional (but valid) details
(75.0\%). Abbreviations and morphological differences also appear to be
important factors. These results underline a problem with the ROUGE
metric, in that it is unable to detect good summaries that use a
different set of words to the reference summary.  One way forward is to
explore metrics that consider sentence semantics beyond word overlap
such as METEOR \cite{banerjee2005meteor} and
\bertscore,\footnote{Indeed, we suggest that \bertscore should be used
  as the canonical evaluation metric for the dataset, but leave
  empirical validation of its superiority for Indonesian summarization
  evaluation to future work.}  and question-answering system
based evaluation such as APES \cite{eyal2019question} and QAGS
\cite{wang2020asking}.
Another way is to create more reference summaries (which will help with the 
issue of the system summaries including [validly] different details to the single reference).

Looking at the results for \textit{average} summaries (middle column),
\bertextabs occasionally fails to capture salient information: 100\% of
the summaries have coverage issues, and 75.0\% contain unnecessary (but valid)
details. They also tend to use paraphrases (87.5\%), which further
impacts on a lower ROUGE score.  Finally, the \textit{bad} system summaries
have similar coverage issues, and also tend to have a very
different focus compared to the reference summary (90.6\%).

In \figref{error} we show two representative examples from
\bertextabs. The first example is considered \textit{good} by our
annotators, but due to abbreviations, morphological differences,
paraphrasing, and additional details compared to the reference summary,
the ROUGE score is $<$0.4.  In this example, the gold summary uses the
abbreviation \textit{kepmenakertrans} while \bertextabs generates the
full phrase \textit{keputusan menteri tenaga kerja dan transmigrasi}
(which is correct).  The example also uses paraphrases (\textit{invites
	strong criticism} to explain \textit{dissatisfaction}), and there are
morphological differences in words such as \textit{tuntutan} (noun) vs.\
\textit{menuntut} (verb).  The low ROUGE score here highlights the fact
that the bigger issue is with ROUGE itself rather than the summary.

The second example is considered to be \textit{bad}, with the following
issues: lack of coverage, wrong focus, and contains unnecessary details
that are not from the article.  The first sentence \textit{President
	Abdurrahman Wahid was absent} has nothing to do with the original
article, creating a different focus (and confusion) in the overall
summary.

To summarize, coverage, focus, and the inclusion of other details are
the main causes of low quality summaries. Our analysis reveals that
abbreviations and paraphrases are another cause of summaries with low
ROUGE scores, but that is an issue with ROUGE rather than the summaries.
Encouragingly, hallucination (generating details not in the original
document) is not a major issue for these models (notwithstanding that
almost 20\% of \textit{bad} samples contain hallucinations).


\section{Related Datasets}
\label{sec:related}

Previous studies on Indonesian text summarization have largely been 
extractive and used small-scale datasets. \newcite{gunawan2017automatic} 
developed an unsupervised summarization model over 3K news articles using 
heuristics such as sentence length, keyword frequency, and title 
features.  In a similar vein, \newcite{najibullah2015indonesian} trained a 
naive Bayes model to extract summary sentences in a 100-article
dataset.  \newcite{aris2012text} and \newcite{silvia2014summarizing} 
apply genetic algorithms to a summarization dataset with less than 200 
articles. These studies do not use ROUGE for evaluation, and the
datasets are not publicly available.

\newcite{koto2016a} released a dataset for chat summarization by manually 
annotating chat logs from 
\textit{WhatsApp}.\footnote{\url{https://www.whatsapp.com/}.}  However, 
this dataset contains only 300 documents. The largest summarization data 
to date is \textit{IndoSum} \cite{kurniawan2018indosum}, which has 
approximately 19K news articles with manually-written summaries.  Based 
on our analysis, however, the summaries of \textit{IndoSum} are highly 
extractive.

Beyond Indonesian, there is only a handful of non-English summarization
datasets that are of sufficient size to train modern deep learning
summarization methods over, including: (1) LCSTS \cite{hu2015lcsts}, which contains
2 million Chinese short texts constructed from the Sina Weibo
microblogging website; and (2) ES-News \cite{gonzalez2019summarization},
which comprises 270k Spanish news articles with summaries.  LCSTS
documents are relatively short (less than 140 Chinese characters), while
ES-News is not publicly available.  Our goal is to create a benchmark
corpus for Indonesian text summarization that is both large scale and
publicly available.

\section{Conclusion}
\label{concl}

We release Liputan6, a large-scale summarization corpus for Indonesian.
Our dataset comes with two test sets: a canonical test set and an
``Xtreme'' variant that is more abstractive. We present results for
several benchmark summarization models, in part based on IndoBERT, a new
pre-trained BERT model for Indonesian. We further conducted extensive
error analysis, as part of which we identified a number of issues with
ROUGE-based evaluation for Indonesian.

\section*{Acknowledgments}
We are grateful to the anonymous reviewers for their
helpful feedback and suggestions. In this research, Fajri Koto is
supported by the Australia Awards Scholarship (AAS), funded by the Department of Foreign Affairs and Trade (DFAT), Australia.
This research was undertaken using the LIEF HPC-GPGPU Facility hosted at The University of Melbourne. This facility was established with the assistance of LIEF Grant LE170100200.

\bibliography{anthology,aacl-ijcnlp2020}

\begin{thebibliography}{36}
\expandafter\ifx\csname natexlab\endcsname\relax\def\natexlab#1{#1}\fi

\bibitem[{{Aristoteles} et~al.(2012){Aristoteles}, {Herdiyeni}, {Ridha}, and
  {Adisantoso}}]{aris2012text}
Aristoteles {Aristoteles}, Yeni {Herdiyeni}, Ahmad {Ridha}, and Julio
  {Adisantoso}. 2012.
\newblock Text feature weighting for summarization of document {B}ahasa
  {I}ndonesia using genetic algorithm.
\newblock \emph{IJCSI International Journal of Computer Science Issues},
  9(1):1--6.

\bibitem[{{Banerjee} and {Lavie}(2005)}]{banerjee2005meteor}
Satanjeev {Banerjee} and Alon {Lavie}. 2005.
\newblock {METEOR}: An automatic metric for {MT} evaluation with improved
  correlation with human judgments.
\newblock In \emph{Proceedings of the ACL Workshop on Intrinsic and Extrinsic
  Evaluation Measures for Machine Translation and/or Summarization}, pages
  65--72.

\bibitem[{{Cheng} and {Lapata}(2016)}]{cheng2016neural}
Jianpeng {Cheng} and Mirella {Lapata}. 2016.
\newblock Neural summarization by extracting sentences and words.
\newblock In \emph{Proceedings of the 54th Annual Meeting of the Association
  for Computational Linguistics (Volume 1: Long Papers)}, volume~1, pages
  484--494.

\bibitem[{{Cohan} et~al.(2018){Cohan}, {Dernoncourt}, {Kim}, {Bui}, {Kim},
  {Chang}, and {Goharian}}]{cohan2018a}
Arman {Cohan}, Franck {Dernoncourt}, Doo~Soon {Kim}, Trung {Bui}, Seokhwan
  {Kim}, Walter {Chang}, and Nazli {Goharian}. 2018.
\newblock A discourse-aware attention model for abstractive summarization of
  long documents.
\newblock In \emph{NAACL HLT 2018: 16th Annual Conference of the North American
  Chapter of the Association for Computational Linguistics: Human Language
  Technologies}, volume~2, pages 615--621.

\bibitem[{{Eyal} et~al.(2019){Eyal}, {Baumel}, and
  {Elhadad}}]{eyal2019question}
Matan {Eyal}, Tal {Baumel}, and Michael {Elhadad}. 2019.
\newblock Question answering as an automatic evaluation metric for news article
  summarization.
\newblock In \emph{NAACL-HLT 2019: Annual Conference of the North American
  Chapter of the Association for Computational Linguistics}, pages 3938--3948.

\bibitem[{{Gehrmann} et~al.(2018){Gehrmann}, {Deng}, and
  {Rush}}]{gehrmann2018bottom}
Sebastian {Gehrmann}, Yuntian {Deng}, and Alexander~M {Rush}. 2018.
\newblock Bottom-up abstractive summarization.
\newblock In \emph{Proceedings of Empirical Methods in Natural Language
  Processing}, pages 4098--4109.

\bibitem[{{Gonzalez} et~al.(2019){Gonzalez}, {Hurtado}, {Segarra},
  {Garcia-Granada}, and {Sanchis}}]{gonzalez2019summarization}
J.-A. {Gonzalez}, L.-F. {Hurtado}, E.~{Segarra}, F.~{Garcia-Granada}, and
  E.~{Sanchis}. 2019.
\newblock Summarization of {S}panish talk shows with siamese hierarchical
  attention networks.
\newblock \emph{Applied Sciences}, 9(18).

\bibitem[{{Graff} et~al.(2003){Graff}, {Kong}, {Chen}, and
  {Maeda}}]{graff2003english}
David {Graff}, Junbo {Kong}, Ke~{Chen}, and Kazuaki {Maeda}. 2003.
\newblock {English Gigaword}.
\newblock Linguistic Data Consortium.

\bibitem[{{Gunawan} et~al.(2017){Gunawan}, {Pasaribu}, {Rahmat}, and
  {Budiarto}}]{gunawan2017automatic}
D~{Gunawan}, A~{Pasaribu}, R~F {Rahmat}, and R~{Budiarto}. 2017.
\newblock Automatic text summarization for {I}ndonesian language using
  {T}ext{T}easer.
\newblock \emph{IOP Conference Series: Materials Science and Engineering},
  190(1):12048.

\bibitem[{{Hermann} et~al.(2015){Hermann}, {Kocisky}, {Grefenstette},
  {Espeholt}, {Kay}, {Suleyman}, and {Blunsom}}]{hermann2015teaching}
Karl~Moritz {Hermann}, Tomas {Kocisky}, Edward {Grefenstette}, Lasse
  {Espeholt}, Will {Kay}, Mustafa {Suleyman}, and Phil {Blunsom}. 2015.
\newblock Teaching machines to read and comprehend.
\newblock \emph{Neural Information Processing Systems}, pages 1693--1701.

\bibitem[{{Hsu} et~al.(2018){Hsu}, {Lin}, {Lee}, {Min}, {Tang}, and
  {Sun}}]{hsu2018unified}
Wan~Ting {Hsu}, Chieh-Kai {Lin}, Ming-Ying {Lee}, Kerui {Min}, Jing {Tang}, and
  Min {Sun}. 2018.
\newblock A unified model for extractive and abstractive summarization using
  inconsistency loss.
\newblock In \emph{Proceedings of the 56th Annual Meeting of the Association
  for Computational Linguistics}, pages 132--141.

\bibitem[{{Hu} et~al.(2015){Hu}, {Chen}, and {Zhu}}]{hu2015lcsts}
Baotian {Hu}, Qingcai {Chen}, and Fangze {Zhu}. 2015.
\newblock {LCSTS}: A large scale {Chinese} short text summarization dataset.
\newblock In \emph{Proceedings of the 2015 Conference on Empirical Methods in
  Natural Language Processing}, pages 1967--1972.

\bibitem[{{Kim} et~al.(2019){Kim}, {Kim}, and {Kim}}]{kim2019abstractive}
Byeongchang {Kim}, Hyunwoo {Kim}, and Gunhee {Kim}. 2019.
\newblock Abstractive summarization of {Reddit} posts with multi-level memory
  networks.
\newblock In \emph{NAACL-HLT 2019: Annual Conference of the North American
  Chapter of the Association for Computational Linguistics}, pages 2519--2531.

\bibitem[{{Koto}(2016)}]{koto2016a}
Fajri {Koto}. 2016.
\newblock A publicly available {I}ndonesian corpora for automatic abstractive
  and extractive chat summarization.
\newblock In \emph{Proceedings of the 10th International Conference on Language
  Resources and Evaluation ({LREC} 2016)}.

\bibitem[{{Koto} et~al.(to appear){Koto}, {Rahimi}, {Lau}, and
  {Baldwin}}]{koto2020indolem}
Fajri {Koto}, Afshin {Rahimi}, Jey~Han {Lau}, and Timothy {Baldwin}. to appear.
\newblock {I}ndo{LEM} and {I}ndo{BERT}: A benchmark dataset and pre-trained
  language model for {I}ndonesian {NLP}.
\newblock In \emph{Proceedings of the 28th International Conference on
  Computational Linguistics (COLING 2020)}.

\bibitem[{{Kurniawan} and {Louvan}(2018)}]{kurniawan2018indosum}
Kemal {Kurniawan} and Samuel {Louvan}. 2018.
\newblock Indosum: A new benchmark dataset for {I}ndonesian text summarization.
\newblock In \emph{2018 International Conference on Asian Language Processing
  (IALP)}, pages 215--220.

\bibitem[{{Lewis} et~al.(2020){Lewis}, {Liu}, {Goyal}, {Ghazvininejad},
  {Mohamed}, {Levy}, {Stoyanov}, and {Zettlemoyer}}]{lewis2019bart}
Mike {Lewis}, Yinhan {Liu}, Naman {Goyal}, Marjan {Ghazvininejad}, Abdelrahman
  {Mohamed}, Omer {Levy}, Veselin {Stoyanov}, and Luke {Zettlemoyer}. 2020.
\newblock {BART}: Denoising sequence-to-sequence pre-training for natural
  language generation, translation, and comprehension.
\newblock In \emph{Proceedings of the 58th Annual Meeting of the Association
  for Computational Linguistics}, pages 7871--7880.

\bibitem[{{Lin}(2004)}]{lin2004rouge}
Chin-Yew {Lin}. 2004.
\newblock {ROUGE}: A package for automatic evaluation of summaries.
\newblock In \emph{Text Summarization Branches Out: Proceedings of the ACL-04
  Workshop}, pages 74--81.

\bibitem[{{Liu} and {Lapata}(2019)}]{liu2019text}
Yang {Liu} and Mirella {Lapata}. 2019.
\newblock Text summarization with pretrained encoders.
\newblock In \emph{2019 Conference on Empirical Methods in Natural Language
  Processing}, pages 3728--3738.

\bibitem[{{Medved} and {Suchomel}(2017)}]{medved2019indonesian}
Marek {Medved} and V\'{i}t {Suchomel}. 2017.
\newblock Indonesian web corpus (id{W}ac).
\newblock In \emph{LINDAT/CLARIN digital library at the Institute of Formal and
  Applied Linguistics (ÚFAL), Faculty of Mathematics and Physics, Charles
  University}.

\bibitem[{{Najibullah}(2015)}]{najibullah2015indonesian}
Ahmad {Najibullah}. 2015.
\newblock Indonesian text summarization based on naive {Bayes} method.
\newblock \emph{Proceeding Of The International Seminar and Conference 2015},
  1(1).

\bibitem[{{Nallapati} et~al.(2016{\natexlab{a}}){Nallapati}, {Zhai}, and
  {Zhou}}]{nallapati2016summarunner}
Ramesh {Nallapati}, Feifei {Zhai}, and Bowen {Zhou}. 2016{\natexlab{a}}.
\newblock Summa{R}u{NN}er: A recurrent neural network based sequence model for
  extractive summarization of documents.
\newblock In \emph{Proceedings of the Thirtieth AAAI Conference on Artificial
  Intelligence (AAAI-16)}, pages 3075--3081.

\bibitem[{{Nallapati} et~al.(2016{\natexlab{b}}){Nallapati}, {Zhou}, dos
  {santos}, {Gulcehre}, and {Xiang}}]{nallapati2016abstractive}
Ramesh {Nallapati}, Bowen {Zhou}, Cicero~Nogueira dos {santos}, Caglar
  {Gulcehre}, and Bing {Xiang}. 2016{\natexlab{b}}.
\newblock Abstractive text summarization using sequence-to-sequence {RNN}s and
  beyond.
\newblock In \emph{Proceedings of The 20th SIGNLL Conference on Computational
  Natural Language Learning}, pages 280--290.

\bibitem[{{Narayan} et~al.(2018){Narayan}, {Cohen}, and
  {Lapata}}]{narayan2018don}
Shashi {Narayan}, Shay~B. {Cohen}, and Mirella {Lapata}. 2018.
\newblock Don't give me the details, just the summary! {Topic}-aware
  convolutional neural networks for extreme summarization.
\newblock In \emph{EMNLP 2018: 2018 Conference on Empirical Methods in Natural
  Language Processing}, pages 1797--1807.

\bibitem[{{Paulus} et~al.(2018){Paulus}, {Xiong}, and {Socher}}]{paulus2017a}
Romain {Paulus}, Caiming {Xiong}, and Richard {Socher}. 2018.
\newblock A deep reinforced model for abstractive summarization.
\newblock In \emph{Proceedings of the 6th International Conference on Learning
  Representations}.

\bibitem[{{Rush} et~al.(2015){Rush}, {Chopra}, and {Weston}}]{rush2015neural}
Alexander~M. {Rush}, Sumit {Chopra}, and Jason {Weston}. 2015.
\newblock A neural attention model for abstractive sentence summarization.
\newblock In \emph{Proceedings of Empirical Methods in Natural Language
  Processing}, pages 379--389.

\bibitem[{{Sandhaus}(2008)}]{sandhaus2008the}
Evan {Sandhaus}. 2008.
\newblock The {N}ew {Y}ork {T}imes annotated corpus.
\newblock Linguistic Data Consortium.

\bibitem[{{See} et~al.(2017){See}, {Liu}, and {Manning}}]{see2017get}
Abigail {See}, Peter~J. {Liu}, and Christopher~D. {Manning}. 2017.
\newblock Get to the point: Summarization with pointer-generator networks.
\newblock In \emph{Proceedings of the 55th Annual Meeting of the Association
  for Computational Linguistics}, pages 1073--1083.

\bibitem[{{Sharma} et~al.(2019){Sharma}, {Li}, and
  {Wang}}]{sharma2019bigpatent}
Eva {Sharma}, Chen {Li}, and Lu~{Wang}. 2019.
\newblock {BIGPATENT}: A large-scale dataset for abstractive and coherent
  summarization.
\newblock In \emph{ACL 2019: The 57th Annual Meeting of the Association for
  Computational Linguistics}, pages 2204--2213.

\bibitem[{{Silvia} et~al.(2014){Silvia}, {Rukmana}, {Aprilia}, {Suhartono},
  {Wongso}, and {Meiliana}}]{silvia2014summarizing}
{Silvia}, Pitri {Rukmana}, Vivi~Regina {Aprilia}, Derwin {Suhartono}, Rini
  {Wongso}, and {Meiliana}. 2014.
\newblock Summarizing text for {I}ndonesian language by using latent
  {Dirichlet} allocation and genetic algorithm.
\newblock In \emph{1st International Conference on Electrical Engineering,
  Computer Science and Informatics 2014}, pages 148--153.

\bibitem[{{Tala} et~al.(2003){Tala}, {Kamps}, {Müller}, and
  de~{Rijke}}]{tala2003the}
F.~{Tala}, J.~{Kamps}, K.E. {Müller}, and M.~de~{Rijke}. 2003.
\newblock The impact of stemming on information retrieval in {B}ahasa
  {I}ndonesia.
\newblock In \emph{The 14th Meeting of Computational Linguistics in the
  Netherlands}.

\bibitem[{{Tan} et~al.(2017){Tan}, {Wan}, and {Xiao}}]{tan2017abstractive}
Jiwei {Tan}, Xiaojun {Wan}, and Jianguo {Xiao}. 2017.
\newblock Abstractive document summarization with a graph-based attentional
  neural model.
\newblock In \emph{Proceedings of the 55th Annual Meeting of the Association
  for Computational Linguistics (Volume 1: Long Papers)}, volume~1, pages
  1171--1181.

\bibitem[{{Wang} et~al.(2020){Wang}, {Cho}, and {Lewis}}]{wang2020asking}
Alex {Wang}, Kyunghyun {Cho}, and Mike {Lewis}. 2020.
\newblock Asking and answering questions to evaluate the factual consistency of
  summaries.
\newblock In \emph{Proceedings of the 58th Annual Meeting of the Association
  for Computational Linguistics}, pages 5008--5020.

\bibitem[{{Wu} et~al.(2016){Wu}, {Schuster}, {Chen}, {Le}, {Norouzi},
  {Macherey}, {Krikun}, {Cao}, {Gao}, {Macherey}, {Klingner}, {Shah},
  {Johnson}, {Liu}, Łukasz {Kaiser}, {Gouws}, {Kato}, {Kudo}, {Kazawa},
  {Stevens}, {Kurian}, {Patil}, {Wang}, {Young}, {Smith}, {Riesa}, {Rudnick},
  {Vinyals}, {Corrado}, {Hughes}, and {Dean}}]{wu2016google}
Yonghui {Wu}, Mike {Schuster}, Zhifeng {Chen}, Quoc~V. {Le}, Mohammad
  {Norouzi}, Wolfgang {Macherey}, Maxim {Krikun}, Yuan {Cao}, Qin {Gao}, Klaus
  {Macherey}, Jeff {Klingner}, Apurva {Shah}, Melvin {Johnson}, Xiaobing {Liu},
  Łukasz {Kaiser}, Stephan {Gouws}, Yoshikiyo {Kato}, Taku {Kudo}, Hideto
  {Kazawa}, Keith {Stevens}, George {Kurian}, Nishant {Patil}, Wei {Wang},
  Cliff {Young}, Jason {Smith}, Jason {Riesa}, Alex {Rudnick}, Oriol {Vinyals},
  Greg {Corrado}, Macduff {Hughes}, and Jeffrey {Dean}. 2016.
\newblock Google's neural machine translation system: Bridging the gap between
  human and machine translation.
\newblock \emph{arXiv preprint arXiv:1609.08144}.

\bibitem[{{Zhang} et~al.(2020{\natexlab{a}}){Zhang}, {Zhao}, {Saleh}, and
  {Liu}}]{zhang2019pegasus}
Jingqing {Zhang}, Yao {Zhao}, Mohammad {Saleh}, and Peter {Liu}.
  2020{\natexlab{a}}.
\newblock {PEGASUS}: Pre-training with extracted gap-sentences for abstractive
  summarization.
\newblock In \emph{ICML 2020: 37th International Conference on Machine
  Learning}.

\bibitem[{{Zhang} et~al.(2020{\natexlab{b}}){Zhang}, {Kishore}, {Wu},
  {Weinberger}, and {Artzi}}]{zhang2020bertscore}
Tianyi {Zhang}, Varsha {Kishore}, Felix {Wu}, Kilian~Q. {Weinberger}, and Yoav
  {Artzi}. 2020{\natexlab{b}}.
\newblock {BERTScore}: Evaluating text generation with {BERT}.
\newblock In \emph{ICLR 2020: Eighth International Conference on Learning
  Representations}.

\end{thebibliography}
\bibliographystyle{acl_natbib}

\appendix

\end{document}